# Document Decomposition of Bangla Printed Text

Md. Fahad Hasan[1], Tasmin Afroz[1], Sabir Ismail[2], Md. Saiful Islam[2]

[1] Department of Computer Science and Engineering, Shahjalal University of Science & Technology, Sylhet, Bangladesh.

[2] Asst. Professor, Department of Computer Science and Engineering, Shahjalal University of Science & Technology, Sylhet, Bangladesh.

**Abstract**: Today all kind of information is getting digitized and along with all this digitization, the huge archive of various kinds of documents is being digitized too. We know that, Optical Character Recognition is the method through which, newspapers and other paper documents convert into digital resources. But, it is a fact that this method works on texts only. As a result, if we try to process any document which contains non-textual zones, then we will get garbage texts as output. That is why; in order to digitize documents properly they should be pre-processed carefully. And while preprocessing, segmenting document in different regions according to the category properly is most important. But, the Optical Character Recognition processes available for Bangla language have no such algorithm that can categorize a newspaper/book page fully. So we worked to decompose a document into its several parts like headlines, sub headlines, columns, images etc. And if the input is skewed and rotated, then the input was also de-skewed and de-rotated. To decompose any Bangla document we found out the edges of the input image. Then we find out the horizontal and vertical area of every pixel where it lies in. Later on the input image was cut according to these areas. Then we pick each and every sub image and found out their height-width ratio, line height. Then according to these values the sub images were categorized. To de skew the image we found out the skew angle and de skewed the image according to this angle. To de-rotate the image we used the line height, matra line, pixel ratio of matra line.

## 1. INTRODUCTION

Today all kind of information is getting digitized and along with all this digitization, the huge archive various kinds of documents are being digitized too. It is known that, optical character recognition (OCR) is the main medium through which any document is transformed into digital data. Now it is a fact that, OCR can process texts only and for non-textual input, it will give garbage text output. So, if we want OCR to work perfectly is must be ensured that it is not fed any kind of non-textual input. Therefore, with a view to transforming document archives into digital collections smoothly and properly, documents are to be sectioned and decomposed category wise. Newspaper page has a complex layout. It contains images along with texts. Again, here the textual zones are divided in various parts like headlines, sub-headlines, columns. Though a few works on decomposing Bangla newspaper/document was done but, these approaches works only if the headline is on the top of the columns, most importantly none of them has handled the situation where a document contains images. So keeping all these in mind we proceeded with a bunch of algorithms, to decompose Bangla newspaper/documents section wise properly. In our process, at first we found the edge of the input image by

"Canny" [3] algorithm. Then we took every pixel of this edge detected image and found out the horizontal and vertical area where it resides. Then converted the input in such a way that, the entire textual zone and image portion became black, and non-textual zones became white. Actually we had to categorize these black portions. This paper is sectioned in following chapters. Chapter 2 sums up some related work on document decomposition and their accuracy. Chapter 3 describes our document decomposition and recognition system, chapter 4 contains our result analysis, and chapter 5 focuses on the future plans and conclusion. For our work, we have used the scanned images collected from some Bangla newspapers.

## 2. RELATED WORKS

Hadjar et al. (2001) worked on newspaper page decomposition [1]. They implemented a split and merge approach.They decomposed English newspapers.

Gatos et al. (1999) also worked on newspaper decomposition and article tracing. Recall and precision of their tracing texts and images were about 96% [2]. They moved ahead with a rule based approach to identify the segments of document.

Omee et al. (2012) worked on decomposing Bangla newspaper. [4] But, their approach works if and only if, there is no image in the input and the headline is on the top of the input.

Bansal et al. also worked on newspaper article extraction. They used hierarchical fixed point model. The accuracy of their labeling headline, sub headline and text blocks form English newspapers are about 96%, 82% and 97% respectively[5].

Ha et al. (1995) worked on document page decomposition. They applied bounding box projection technique [6].

Gao et al. (2007) also decomposed document[7]. They implemented integer linear programming. They did not handle images of the input.

## 3. DOCUMENT DECOMPOSITION and RECOGNITION

Document decomposition is a process where various regions of any scanned image or scanned text document are identified and categorized. Its main aim is to distinguish the textual zone from non-textual zone. Documents' layout may vary depending on the sources (newspaper, text books, magazines etc.) they have been taken from. The key parts of any document are identified through document decomposition. Like for a newspaper key parts are images, headlines, columns, sub-headlines.

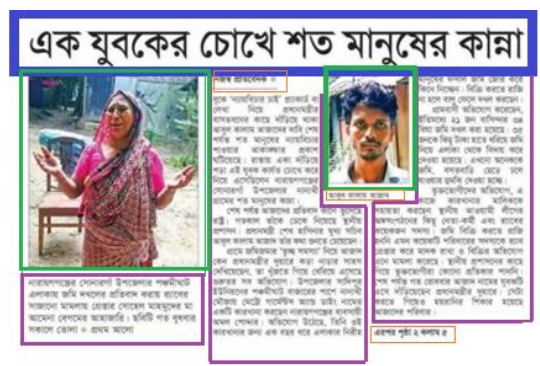

Fig.1 Various regions of a document

For document decomposition, at first we analysed the layout of taken input. We all know that each page maintains a certain gap between the margins and the textual zone. We used this characteristic of pages to find out the area from where the main document is started. Then we detected the edges of each element. Next, according to some threshold values we converted the edge detected image into an image which contains only white and black pixels. Then we distinguished elements separators. Later, we recognized the elements.

The images of any document occupy a certain ratio of height and width, a specific pixel ratio, as well as a particular ratio of the sum of the total length and total width. We specified these characteristics to determine images. When it comes to text zones, we know that the descending order of the height of the letters of any document are headline, sub headline, column. We implemented methodology to calculate the letter height and then recognized the textual zones. Next we worked on de-skewing and de-rotating. A document may start with textual zones or image. Our process worked properly to de-skew and de-rotate whether they start with text or image.

All our methods are abridged below.

### 3.1 Edge Detection

At first, we took an input image in form of matrix. Then we found out the edges in the input image and mark them in the output map using the 'Canny' algorithm. Before applying this algorithm, it was ensured that, the image and the edges of various portions of the image remains noiseless and fairly sharp respectively. Fig.2 shows the input image on left side and the edge detected image on right side.

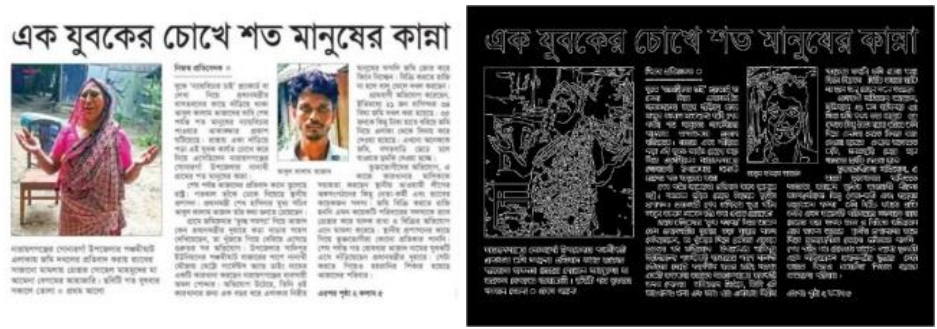

Fig.2   Input image and edge detected image

### 3.2 Conversion of Input Image

In this step, we pick every pixel of the input and assigned them a value according to the length of the vertical and horizontal line it belongs to. That is every pixel of same line will be given a same value. Here we assigned 0 (black) for pixel to the array if the pixel color is not black (>0). Thus we will get an image where, the textual area will be black and the not textual area will be white. The converted image for the Fig.2 is shown in Fig.3.

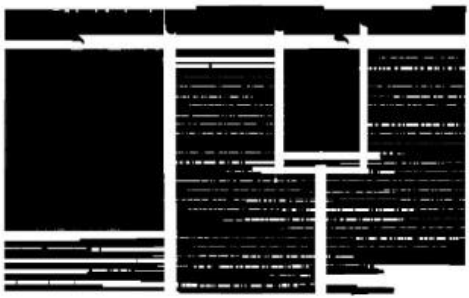

Fig.3   Converted image

## 3.3 Discerning the Element Separators

Now, we stored the information of every pixel from the converted image. Like 1 for white colour and 0 for black colour. We picked pixels from the converted image and compared the value of height and width, they are assigned with, with some threshold values to determine whether they belong to non-textual area or not.

## 3.4 Segmenting Black Boxes

In this stage, we stored the top-left and bottom-right coordinates of each black box, neighbouring the horizontal or vertical element separators of the converted image. Then we cut our input according to the coordinates of the black boxes. This cut blocks are send to detect their category.

## 3.5 Distinguishing the Elements

### 3.5.1 Recognizing Images

We proposed three filters to define a region as image. First filter checks, whether the length and width of each block is greater than a threshold value or not. Second one finds the pixel ratio of the block and compares them with a threshold value. Third filter determines the ratio of the sum of the total length of each block and total width of each block. The detected images of Fig 2 are shown here.

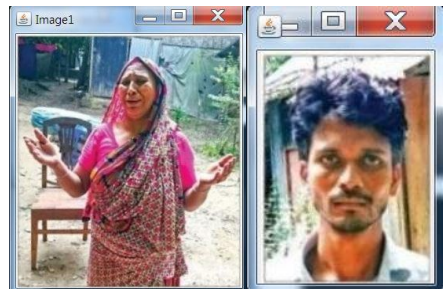

Fig.4   Detected images

### 3.5.2 Labelling the Text Zones

To categorise the regions of text zones into headline, sub headline and columns we moved ahead with the methodology of dominant line/letter height. We checked out the line heights of inputs from several 'epapers'. We found out the gap between height of dominant letters, height of headlines, and sub headlines. To detect the letter heights perfectly, we removed the image portion from the input after detection. Through those results, we decided some threshold value 'gap1', 'gap2', 'x1', 'x2', 'x3' to label the text regions.

If, Line height difference> gap1 && line height>=x3, Then it is labelled as headline.
If, Line height difference> gap2 && x1<line height<x2, Then it is decided to be sub headline.
Other text zones are columns.
In Fig.5, Fig.6 the detected headlines and columns are shown.

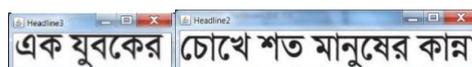

Fig.5   Detected headline

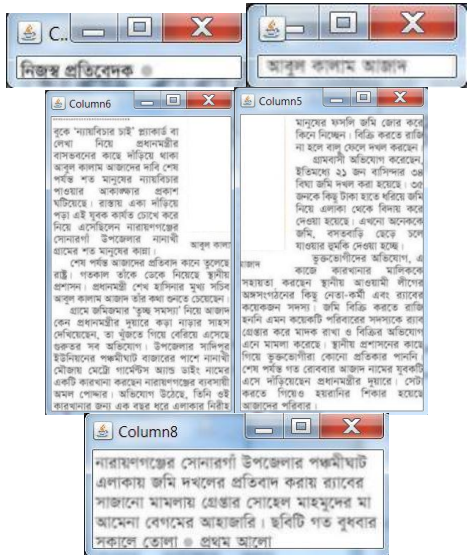

Fig.6   Detected colums

### 3.6 Image Auto Rotation and De-skew

Input may be both skewed and rotated. To solve this problem, we computed the skew angle of the input. Input may contain an image portion at the beginning. For such cases to compute the skew angle perfectly the input was rotated to 90º, 180º and -90º. Then skew angles of all these rotated inputs were calculated. We picked the highest absolute value of skew angles and de-skewed the input according to this value. After this the de-skewed image was auto rotated. For this we calculated the line height, matra line, index point, total black pixel on matra line for both the input image and 90º rotated image. Then we compared the data computed from previous step. We compared the pixel ratio (pixel/line height), and took the bigger value. Now, we checked whether 'the half of the line height' is greater than 'the matra line index' or not. In addition, depending on this we decided whether we have to rotate the image or not. If rotation is needed, we will return 180º or -90º, depending on which pixel ratio value we chose.

When the input contains a large image portion at the beginning then, in order to ensure the flawless performance of de-skew and rotation, the line height, matra line, index point, total black pixel on matra line is also calculated for the last line of the input, and 90º rotated image. In Fig 7, you can see a skewed and rotated input containing image at the beginning, and the de-skewed and de rotated output of it.

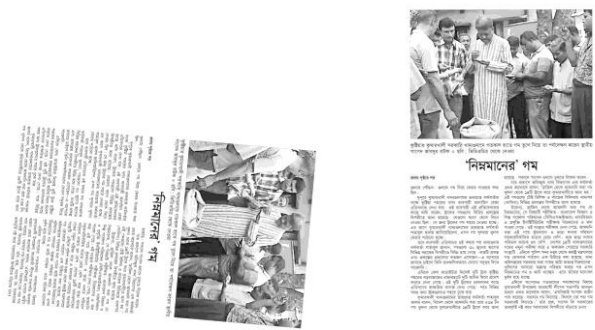

Fig.7   Skewed and rotated input, de-skewed and de-rotated output

## 4. RESULT ANALYSIS

We checked about 70 skewed images. The accuracy of our de- skew system is approximately 100%. We checked 70 skewed as well as rotated images. Then the accuracy was 97.83%.
We decomposed about 300 scanned newspaper pages. The precision, accuracy and recall of our methodology are given below.

Table 1 Result Analysis Table

| Elements | Precision | Recall | Accuracy |
|---|---|---|---|
| Images | 80.70% | 90.0% | 98.35% |
| Headlines | 93.04% | 91.45% | 97.07% |
| Sub-headlines | 83.15% | 88.76% | 96.75% |
| Columns | 97.22% | 93.47% | 95.89% |

## 5. CONCLUSION

Documents may also contain table. Still we did not find out the features for tables. So, at this moment if any document contains table, then it is detected as image. So, in future, tables should be detected. And if the noise and images can be detected and removed before de-skew operation then the output will be 100% accurate. For 100% flawless decomposition and recognition, the noise from the input image must be removed.